\documentclass[conference]{IEEEtran}
\IEEEoverridecommandlockouts
\usepackage{cite}
\usepackage{amsmath,amssymb,amsfonts}
\usepackage{algorithmic}
\usepackage{graphicx}
\usepackage{textcomp}
\usepackage{xcolor}
\usepackage{url}
\usepackage{enumitem}
\def\BibTeX{{\rm B\kern-.05em{\sc i\kern-.025em b}\kern-.08em
    T\kern-.1667em\lower.7ex\hbox{E}\kern-.125emX}}
\begin{document}

\title{On the Behavior of Audio-Visual Fusion Architectures in Identity Verification Tasks
}

\author{
\IEEEauthorblockN{{Daniel Claborne}\IEEEauthorrefmark{1}}
\and
\IEEEauthorblockN{{Eric Slyman}\IEEEauthorrefmark{2}}
\and
\IEEEauthorblockN{{Karl Pazdernik}\IEEEauthorrefmark{1}\IEEEauthorrefmark{3}}\newline
\and
\IEEEauthorblockA{\IEEEauthorrefmark{1} \textit{Pacific Northwest National Laboratory}}
\IEEEauthorblockA{\IEEEauthorrefmark{2} \textit{Oregon State University}}
\IEEEauthorblockA{\IEEEauthorrefmark{3} \textit{North Carolina State University}}
}

\maketitle

\section{Abstract}

We train an identity verification architecture and evaluate modifications to the part of the model that combines audio and visual representations -- including in scenarios where one input is missing in either of two examples to be compared.  We report results on the Voxceleb1-E test set that suggest averaging the output embeddings improves error rate in the full-modality setting and when a single modality is missing, and makes more complete use of the embedding space than systems which use shared layers and discuss possible reasons for this behavior.

\section{Introduction}

Identify verification is a common task in audio and visual analysis.  Neural networks of various architectures have shown great success in matching audio and visual signatures of the same entity.  More recently, work has been done to augment the ability of these systems to perform identify verification by combining information from both the audio and visual streams.  A more difficult task for audio-visual fusion for identify verification is to perform verification in situations when one or both examples in a pair to be matched are missing a modality that is present in the other example.  Systems that attempt this task must be able to form a representation from each modality that is useful even if the other modality is missing while still performing well when both modalities are present. 

In this work we refer to comparing audio-visual examples of two entities in the scenarios:  \textit{cross-modal}, where the audio from one example is compared to the video in the other, \textit{mixed-modal}, where one of the two examples is missing a single modality, \textit{full-modality}, where both modalities are present in both examples, and \textit{uni-modal}, where only one modality is present in both examples. 

\section{Previous Work}

\subsection{Uni-Modal Models}
Identity verification and entity similarity are widely studied in the uni-modal setting, being used to recognize faces, voices, and similar images \cite{wan_generalized_2020}\cite{wang_speaker_2018}\cite{fan_exploring_2021}\cite{mehdipour_ghazi_comprehensive_2016}.  Training for these networks usually takes the form of trying to pack examples in the same category together in an embedding space while moving them away from examples in different categories.  Loss functions that encourage this behavior in training include generalized end-to-end loss \cite{wan_generalized_2020} and arc-margin loss \cite{deng_arcface_2019}.  Both audio and visual modalities often rely on convolutional neural network architectures to form embeddings.  Convolutional neural networks act upon audio either through temporal convolutions over the raw audio signal \cite{baevski_wav2vec_2020}, or more commonly through two dimensional convolutions over spectrogram maps of the audio.  Transformer architectures have also been used recently in audio recognition to exploit the time-series nature of the modality \cite{fan_exploring_2021}\cite{liu_mockingjay_2020}.

\subsection{Multi-modal Models}
Models for audio-visual tasks are usually represented by two uni-modal networks each taking in one of the audio or visual modalities and then combining (fusing) features at some point in the network; there is also then a choice of how the features are combined.  This setup is used in tasks such as audio-visual alignment \cite{chung_out_2017}, identity verification \cite{shon_noise-tolerant_2019}\cite{chen_multi-modality_2020}, and video diarization \cite{kang_multimodal_2020}\cite{chung_who_2019}.  

The most common location to perform the fusion is near the output, since it is easy to force each modality to a particular dimension that makes fusion straightforward.  Where the dimensions of the two input streams can be reasonably aligned, features can be fused at some intermediate layer such as in dual networks taking in raw video and optical flow \cite{feichtenhofer_convolutional_2016}\cite{carreira_quo_2018}.  The attention-based architecture of \cite{jaegle_perceiver_2021} showed success in combining the two modalities at the input with a padding scheme.  Schemes for fusing the embeddings between the two modalities include concatenation, averaging, learning a combination through a linear layer, LSTM layers, or self-attention \cite{afouras_deep_2018}\cite{petridis_end--end_2017}\cite{sari_multi-view_2021}.  The architecture of \cite{sari_multi-view_2021} uses a shared embedding layer that processes the output of both the audio and visual modalities and optimizes a joint loss function.

Generally, systems which combine multiple modalities improve over those with single modalities, as they have the option of exploiting the extra information and, in some architectures, intelligently combining all modalities \cite{feichtenhofer_convolutional_2016}.  However, this is not always the case and various reasons have been postulated, such as early stopping being triggered by a bad joint solution between modalities \cite{wang_what_2020}.

\section{Model Architecture and Training}

We train several model architectures for the task of identity verification.  To best ensure we are comparing only differences arising from the fusion step, we keep the `backbones' of each model the same.  The audio backbone consists of the MobileNetV2 \cite{sandler_mobilenetv2_2019} architecture with parameters identical to those reported in \cite{sari_multi-view_2021}.  The visual backbone follows a variant of the ResNet convolutional neural network architecture \cite{he_deep_2015} with 4 bottleneck blocks each containing 3 convolutional layers.  During training and inference, it takes in a stack of frames and averages across the time dimension at the output.  The three fusion techniques we investigate are referred to as multi-view, mean fusion, and MLP fusion.  Since the corresponding code and trained model were not available, multi-view fusion is our best recreation of the scheme in \cite{sari_multi-view_2021}, in which each output is passed through a shared embedding layer.  Mean fusion runs each backbone output through a separate linear layer and averages the two outputs.  MLP fusion runs the concatenated output of both backbones through a multi-layer perceptron.  For the multi-view model fusion, we employ the same linear layer that processes the output of both backbones.  Details of both backbones and fusion architectures are given the appendix.

\begin{figure*}[ht]
    \centering
    \includegraphics[width=\textwidth]{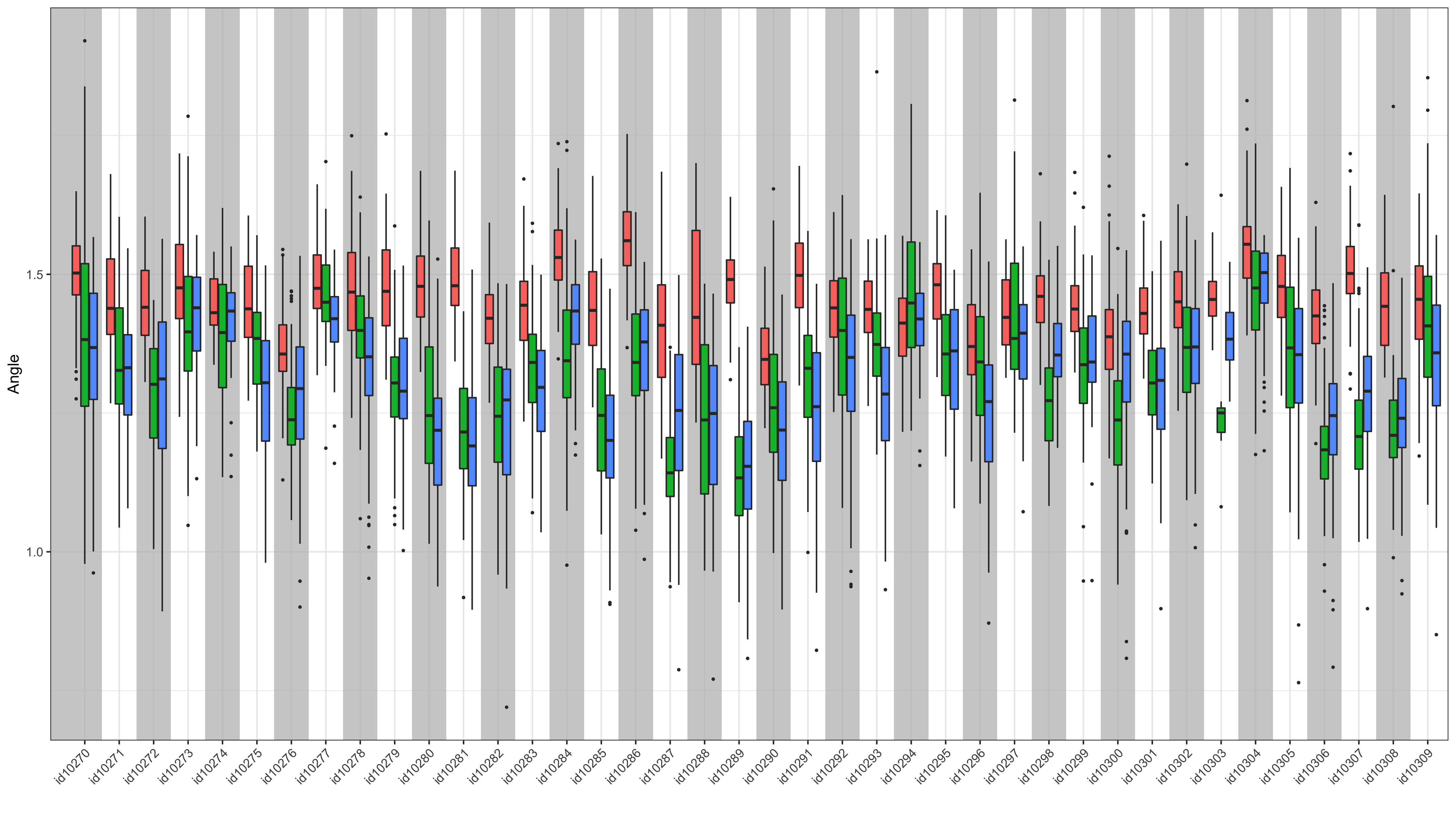}
    \caption{For each identity, a group of boxplots representing the values of angle differences between audio and video embeddings for mean-fusion (Red), MLP (Green), multi-view (Blue).}
    \label{fig:aud_vid_diff}
\end{figure*}

\begin{figure*}[ht]
    \centering
    \includegraphics[width=\textwidth]{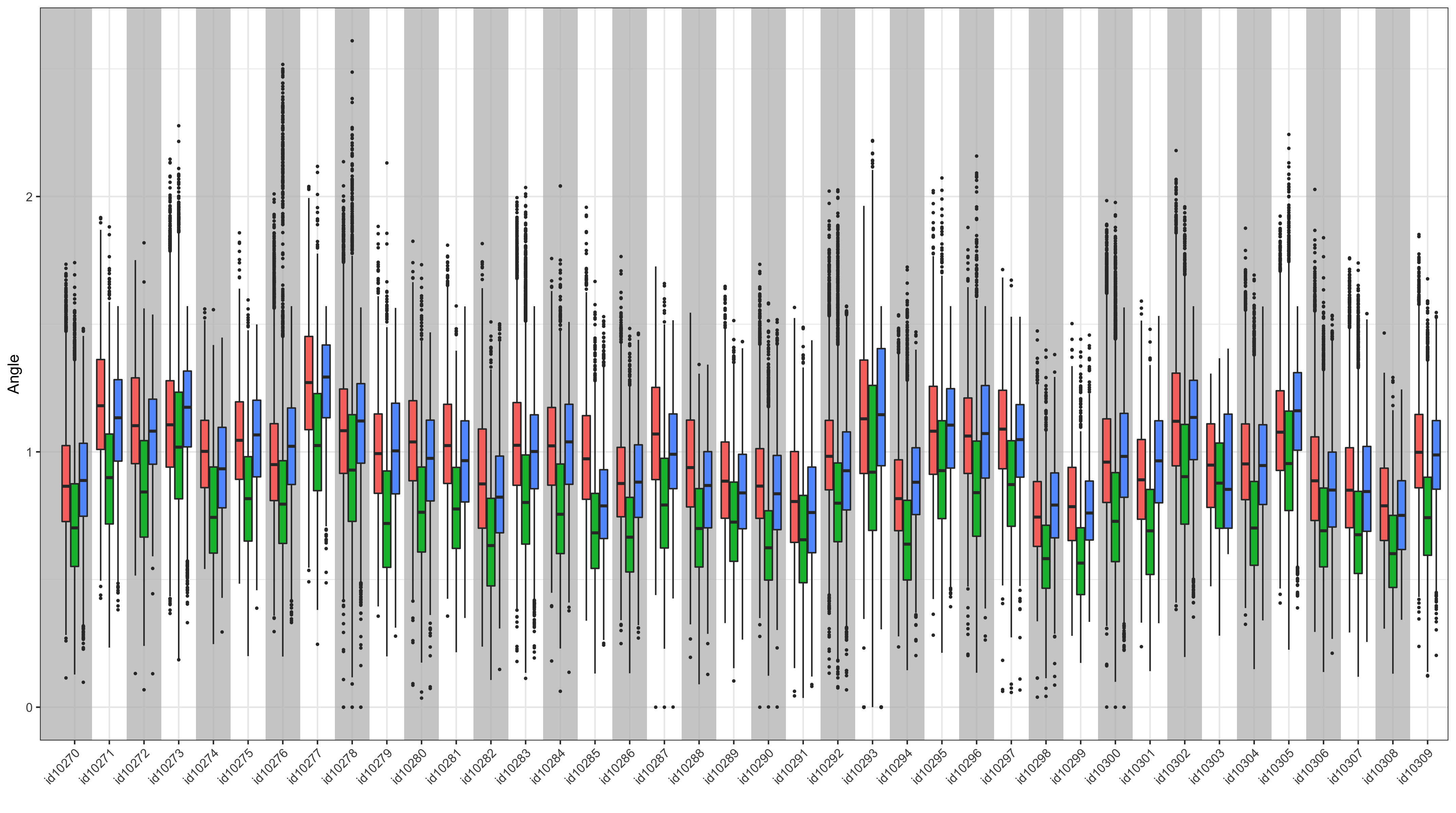}
    \caption{For each identity, a group of boxplots of the of pairwise angle difference between audio embeddings for that identity for each model:  Mean fusion (Red), MLP (Green), multi-view(Blue).}
    \label{fig:pairwise_aud_diff}
\end{figure*}

\begin{figure*}
    \centering
    \includegraphics[width=\textwidth]{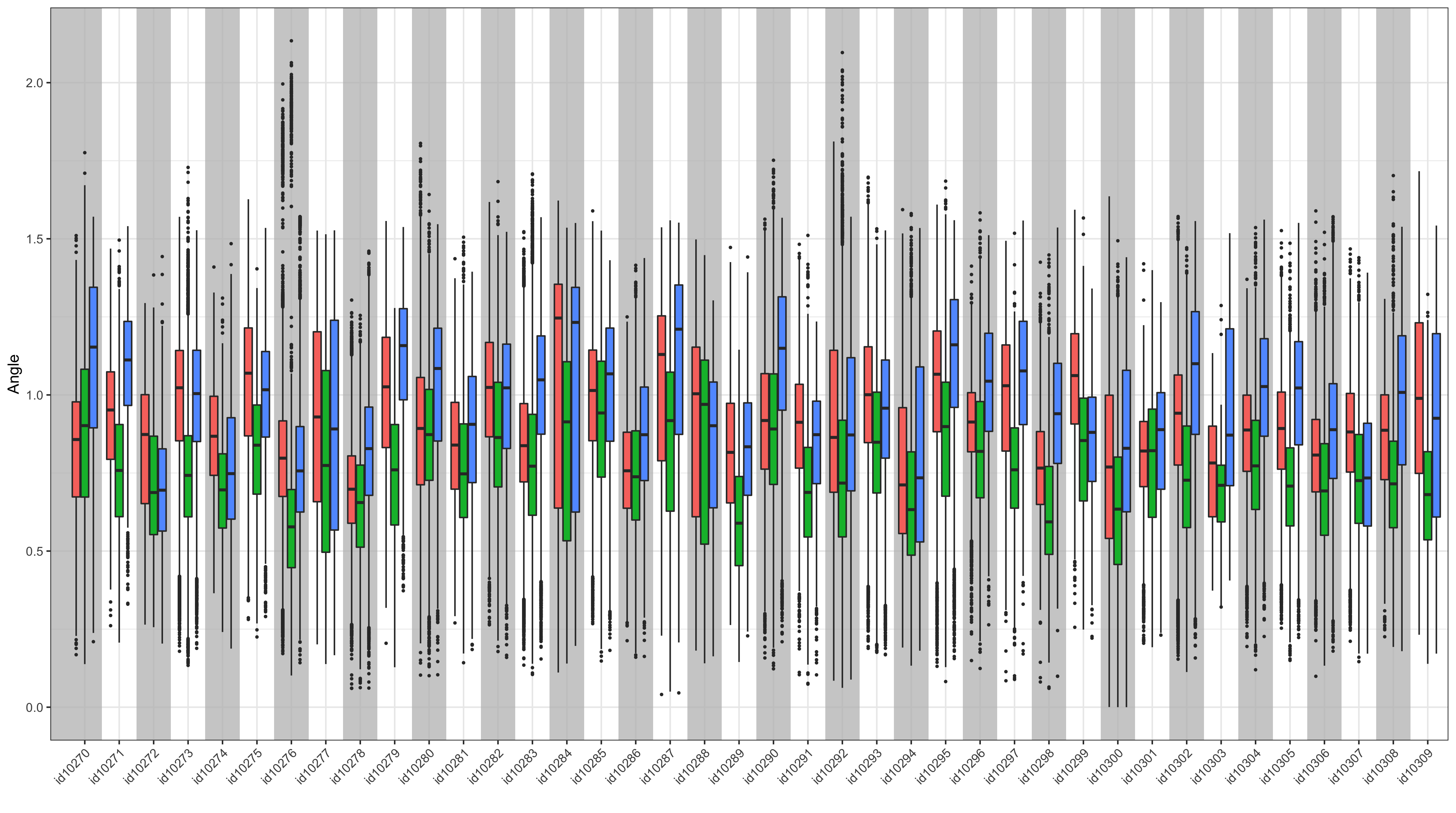}
    \caption{For each identity, a group of boxplots of the of pairwise angle difference between video embeddings for that identity for each model:  Mean fusion (Red), MLP (Green), multi-view (Blue).}
    \label{fig:pairwise_vid_diff}
\end{figure*}

\begin{figure}
    \centering
    \includegraphics[width=\columnwidth]{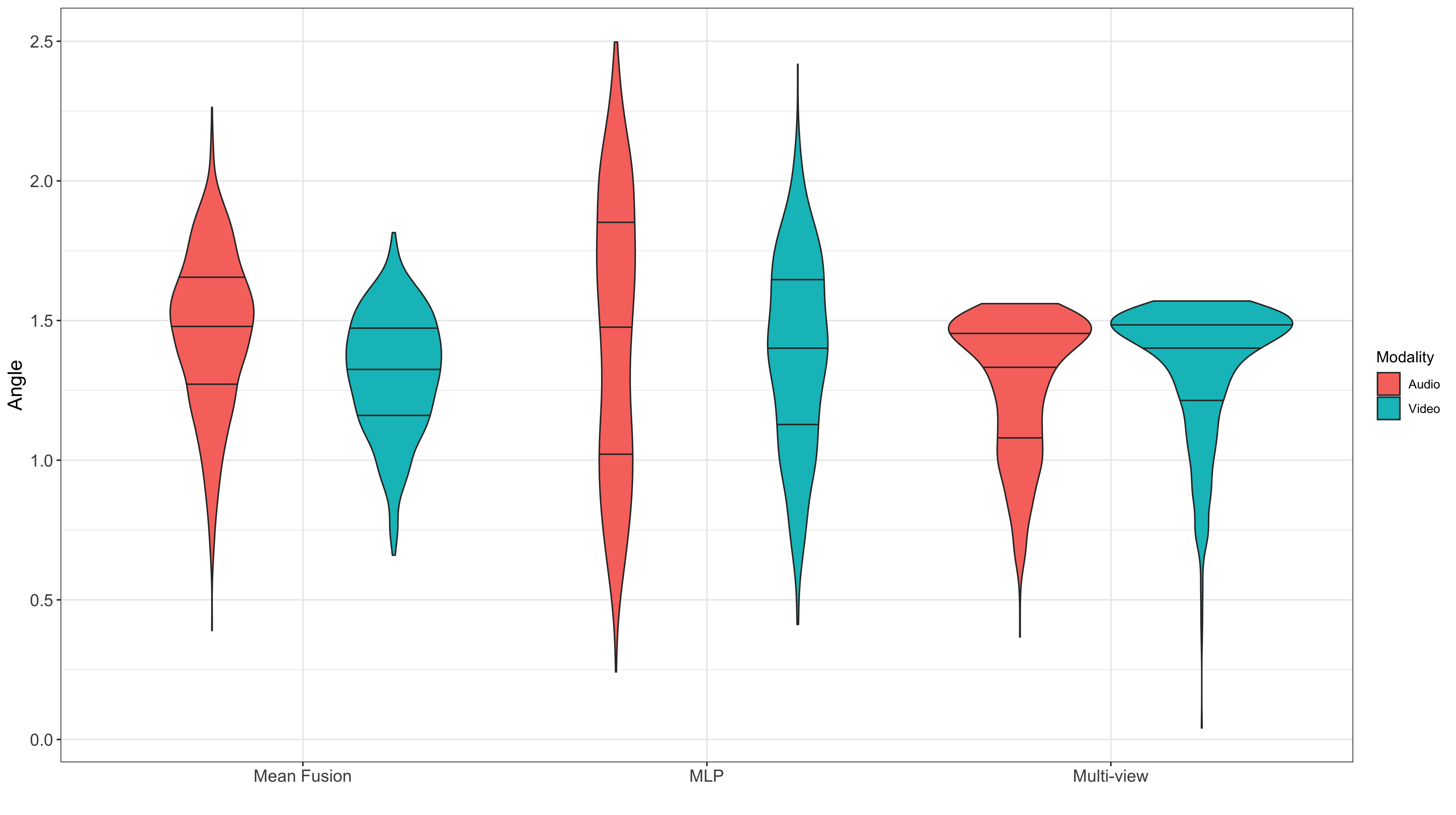}
    \caption{Boxplots of the angles between the embedding centroids of every identity pair, for each model/modality combination.}
    \label{fig:between_id_diffs}
\end{figure}

\begin{figure*}[ht]
    \centering
    \includegraphics[width=\textwidth]{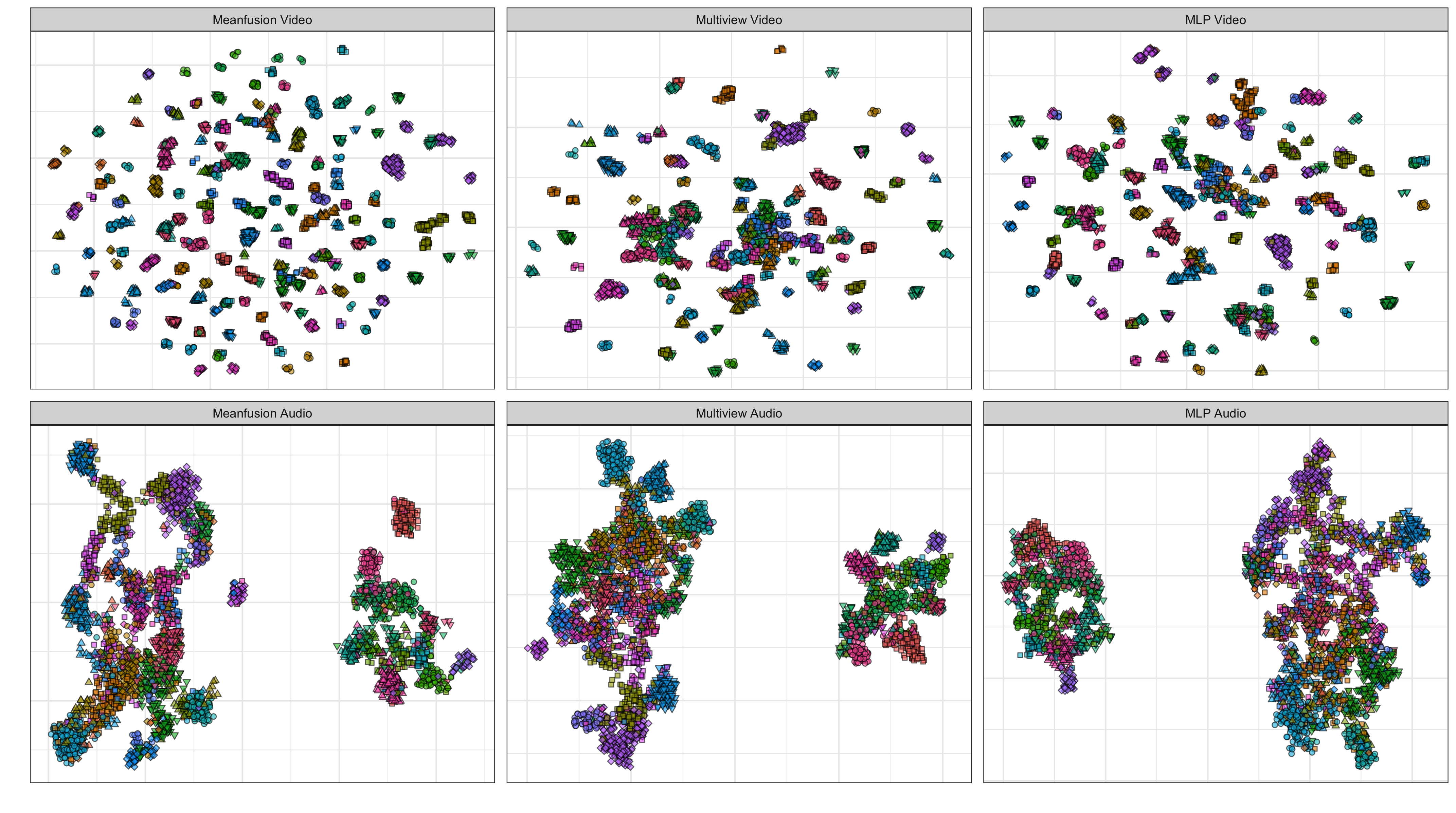}
    \caption{UMAP representations of network embeddings for architecture-modality combinations, colored by speaker identity.  Identities represented by shape/color combinations.  (Columns from left to right:  Mean fusion, multi-view, MLP.   Top/bottom row: video/audio)}
    \label{fig:umap_comparison}
\end{figure*}

We train on the Voxceleb2 \cite{nagrani_voxceleb_2020} train set for up to 10 epochs, using an Adam optimizer with parameters specified in appendix B.  Learning rate is reduced by a factor of 0.95 for every epoch that validation accuracy does not improve.  We use the arc-margin loss from \cite{deng_arcface_2019} with the different classes being the identities in the VoxCeleb2 train set.

\subsection{Cross-Modal Compatibility}
Cross-modal verification is when we are interested in matching entities based on different modalities from those entities, i.e. matching entities A1 and A2 using audio from A1 and video from A2.  Mixed modal verification describes matching entities where one of the two modalities is missing for one of the entities.  All three fusion techniques can be used to perform cross-modal and mixed-modal verification.  
In the multi-view model, the intent is to have cross-modal verification baked into the architecture.  There is a shared classifier $C$ with trainable weights that separately processes the inputs from the audio and visual stream.  The total loss is a weighted combination of the two arc-margin losses produced by each modality:  $$L_{total} = \lambda_1 L(C(x_{a})) + \lambda_2 L(C(x_{v}))$$  Where $L$ is the arc-margin loss, and $x_{a}, x_{v}$ are audio and visual inputs from the same video segments. For our experiments, we set both the $\lambda_1$ and $\lambda_2$ to 0.5.  The arc-margin loss is critical here, as it holds a representation of each speaker that the embedding of both modalities must move towards to minimize the loss.  Ideally, this model will map both the audio and visual embeddings to similar locations in the embedding space during training to satisfy both $L(C(x_{a}))$ and $L(C(x_{v}))$.  Cross-modal verification is then accomplished by passing audio-only from one sample and video-only from another and comparing their embeddings.

To perform cross-modal verification with the mean fusion and MLP-fusion, we train the networks to learn a `null representation', that is, an embedding produced from an input of all zeros.  When the network wants to form an embedding for an input with only one modality, one would simply feed the network for the missing modality an input of all zeros in the appropriate shape, the output representation of which it has learned during training, that output is then combined with the output from the non-missing modality network by average or MLP fusion.  

We hypothesize that for mean fusion, it attempts to learn the average embedding of the missing modality as the null representation during training.  This intuition follows from the behavior of arc margin loss.  Arc-margin maximizes the angle between embeddings in different classes while minimizing the angle for embeddings within the same class.  Suppose we have learned an optimal set of weights for this loss ---no weights can be updated in any manner to improve the loss over the training set--- and we now introduce missing modalities into training.  For a particular class i, in the case of a missing modality, a reasonable solution would be to map the null input to the average of the embeddings for the missing modality for that class.  However, since the other classes have, in this hypothetical scenario, been successfully mapped to different regions of the embedding space, this would come at the expense of training examples with missing modalities for the other classes. Thus, a natural compromise would be to try to satisfy all classes by making the null input map to the average of the embedding centroids for all classes.  For MLP-fusion it is less clear where the network will attempt to place the null-embedding, as its combination with the embedding from the observed input is learned.

\section{Performance}
Though we trained on Voxceleb2, the test set for Voxceleb2 has no agreed upon set of testing pairs, making comparison to other models difficult.  Instead we evaluate the performance of all architectures on the Voxceleb1\cite{nagrani_voxceleb_2017} testing pairs.  We evaluate using the model weights that had the best validation accuracy at the end of an epoch among all 10 training epochs.  We cropped the faces from the source YouTube videos using an implementation \cite{de_paz_centeno_mtcnnlicense_nodate} of the mtCNN architecture \cite{zhang_joint_2016}.  Syncnet \cite{chung_out_2017} was used to select which face-crop was the active speaker, and finally a facial recognition system for the identities in VoxCeleb1 implemented in \cite{malli_keras-vggface_2022} is used to verify that the crops are the correct identity; 18 of 4061 samples from the VC1 test set were not able to be successfully recovered, and are listed in appendix C.

Table \ref{tab:vc1-performance} shows the equal-error-rate (EER) of the three architectures for a verification task on our Voxceleb1 test set in the presence of 5 modality modes:  Full audio-visual (AVxAV), audio only (AxA), video only (VxV), one input with no video (AVxA), one input with no audio (AVxV) and audio-only vs video-only (AxV).  There is no unanimous best model, however, mean fusion is optimal or second best on all ``matched modality'' tasks.  Replicating a result reported in \cite{sari_multi-view_2021}, we actually see weaker scores in the AVxA and AVxV tasks than their uni-modal AxA and VxV counterparts across all models.  This would also seem to agree with the difficulty of training multi-modal systems to improve upon their uni-modal counterparts discussed in \cite{wang_what_2020}, though, we see an opposite trend with better performance in the full audio-visual training regime. Also, if we average across all matched modality scenarios, we notice an overall improved performance in EER when using mean fusion (multiview = 15.4, mean fusion = 14.5, MLP fusion = 15.2).

It is worth noting that these numbers significantly lag performance reported by \cite{sari_multi-view_2021}, which achieve error rates in the low single digits for all evaluation modes except AxV, the only task without a matched modality.  Despite our best efforts to follow the details provided in \cite{sari_multi-view_2021}, various aspects of the training or architecture are undocumented and possibly different, including the choice of optimizer, data augmentation and preprocessing techniques, dropout probability, clip length, and number of training epochs.    Despite this lack of agreement, we believe that investigation into the differences in embedding distributions across models is still valid since all aspects of the architecture, except for the final fusion step, remain the same.

\begin{table}
\caption{Performance of models on the Voxceleb1-E test set}
\centering
    \begin{tabular}{|c|c|c|c|c|c|c|}
        \hline
         & \multicolumn{5}{c|}{Matched} &
        \multicolumn{1}{c|}{Unmatched} \\
         & \multicolumn{5}{c|}{Modality} &
        \multicolumn{1}{c|}{Modality} \\
        \hline
        Model &   AVxAV   &   AxA &   VxV &   AVxA    &   AVxV    & AxV \\
         \hline
         Multi-view  &   10.6 &   \textbf{15.8} &   15.2 &   17.2 &   18.1  & \textbf{29.7}\\
         \hline
         Mean fusion &   \textbf{10.1} &   16.0 & 14.6 &   \textbf{17.0} &   \textbf{15.0} & 35.9 \\
         \hline
         MLP fusion &   10.5 &   17.7 &   \textbf{14.3} &   18.4 &   15.1  & 30.9 \\
         \hline
    \end{tabular}
    \label{tab:vc1-performance}
\end{table}

\section{Embedding Distribution Investigation}

Our experiments are concerned with the difference between embeddings in the single modality case.  For the multi-view model, this is simply the output of passing a single modality through the appropriate backbone.  For the mean fusion and MLP models, this is the result of passing one modality and a null input through the model and extracting the resulting embedding after the fusion step.  We expected that the mean fusion model has more freedom to spread out its embedding representations, as any pair of audio/video embeddings that average to a vector that is close in angle to the target vector will have low loss during training, which could include a pair that occupy very different regions of the embedding space.  In contrast, the multi-view model is constrained in that it must learn to use the same set of weights and activations at the output to process both the audio and visual streams, leading to a solution that favors audio/video embeddings which are more similar to each other.  MLP fusion is also compared as an example of a model that has neither an easy interpretation of how the embeddings are combined (as in mean fusion) nor a training routine that shares weights across two tasks (as for the multi-view model).

\subsection{Angle Between Audio and Video}

For a 3-second clip from each video, we compute the absolute angle between the audio embedding and the video embedding for a given model.  Figure \ref{fig:aud_vid_diff} shows boxplots for each speaker identify of these angles for all videos of that speaker for each of the three models.  We can see that in general, the angle between the audio and video embeddings for the mean fusion model is larger than that of the multi-view or MLP models, which suggests that mean fusion is exploiting its flexibility in the choice of vectors that average to a correct prediction.  This solution does not seem to come at the expense of error rate.  The multi-view system displays the expected behavior of having more similarity between the audio and video embeddings, though not any more than the MLP model.

\subsection{Within and Between Identity Variation}

Recall that the purpose of the arc-margin loss is to tightly cluster the utterances from the same speaker while spreading out clusters of different speakers. Therefore, we are also interested in whether any model has more variance in its embeddings within a speaker and between speakers. To measure the within-speaker variance, for a particular modality, we compute the absolute angle $\theta_{m,k,(i,j)}$ between embeddings $x_{m,k,i}$ and  $x_{m,k,j}$ for model $m$, identity $k$, and sample pair $(i,j)$.  Figures \ref{fig:pairwise_aud_diff} and \ref{fig:pairwise_vid_diff} show boxplots ---for audio and video embeddings, respectively--- of these angle differences for each identity.  These plots do not reveal a significant difference in the within-cluster dispersion of the embeddings between the mean fusion and multi-view models, however both appear to have greater within-cluster dispersion than the MLP model; these plots suggest that the MLP model packs embeddings within a speaker closer together.

To measure the between identity variance, we compute angles between speaker centroids for each model and modality.  Specifically, for a given modality and pair of identities $(i,j)$, we compute the angle difference between the average embedding of each identity. This results in an angle difference in mean embeddings for all speaker pairs.  Figure \ref{fig:between_id_diffs} shows boxplots of these mean differences for all models and modalities. 
The multi-view model shows a truncated distribution, reluctant to place embedding centroids far away from each other in either modality.  The mean fusion model has more cluster centroids with larger angle differences, but not as much as the MLP model, which appears to have a greater range of between-identity differences.

The findings for within and between cluster differences suggests that the MLP model is closest to achieving the objective set by the arc-margin loss, that is, to have speaker clusters evenly distributed on a hypersphere and have their embeddings be tightly packed.  This does not, however, result in better EER on the Voxceleb1-E test set, especially in the comparisons with audio.  It may be that while the MLP model has made more complete use of the embedding space, it has still confused several audio examples, leading to low EER.

We use UMAP \cite{mcinnes_umap_2020} to get a different view of the differences between the embedding spaces for each modality across the three models.  Figure \ref{fig:umap_comparison} shows UMAP embeddings constructed using euclidean distance and n\textunderscore neighbors = 5.  First, and most strikingly, the audio embeddings cluster into two distinct groups for all models in the UMAP space, do not show any appreciable difference in overall structure, and are much less evenly distributed than those of the video embeddings.  This is most likely the result of voice identification being a much more difficult task than facial identification, and perhaps amplified by a tradeoff the network has to make between doing well in the visual task vs the audio task.  Another possibility is that we simply needed to train longer.  Though we saw non-improvement in the validation accuracy for several epochs by the end of training, Wang et al. \cite{wang_what_2020} mention that early stopping (or in our case, using the best validation error model after 10 epochs) leads to underfitting of the RGB stream.  The UMAP embeddings, however, indicate that the RGB stream is well fit, and that perhaps we have failed to find a good \textit{joint} solution that fully utilizes the audio stream.

The video embeddings for the mean fusion model appear much more evenly distributed than those of the multi-view or MLP models.  Evaluating the performance of each clustering with silhouette scores appears to confirm this visual inspection.  The mean fusion clustering achieves the highest silhouette score of 0.258, with multi-view and MLP scoring 0.127 and 0.197.  This apparent difficulty in separating clusters evenly may be the cause of the lower performance of the multi-view method in all of the video tasks.
One possible reason that the multi-view model cannot separate classes as well is due to the limited capacity of the shared classification layer ---the network would like to separate two video classes, but cannot without degrading performance in the more difficult audio task.  The classification layer of multi-view, from the perspective of the visual stream, is in a way regularized by the audio task.  Similarly, the MLP model will have to update the weights in the fusion step to account for the audio inputs, affecting performance even in the case of null audio input.  Mean fusion on the other hand has independent updates (conditional on the multi-modal target embedding), and so the visual network can update itself more freely.

Despite the apparent shortcomings of the multi-view architecture in separating video embeddings, it still performs comparably to the other architectures in the AVxAV and VxV tasks (but falls quite behind in the mixed-modal AVxV task).  For AVxAV, we hypothesize that one modality can `pick up the slack' for any failure in the other modality.  Indeed the experiments in the original implementation of multi-view suggest that this is the case, where they show that score-fusion (combining cosine similarity scores from each modality) is an effective approach to improve EER.  The MLP model does not appear to suffer at all from the apparent limited ability to evenly separate video embeddings, and achieves the lowest error on the VxV task.  Both these results are counter to the silhouette scores and visual inspection of the embeddings in UMAP space where the mean fusion model appears to be more effectively separating examples.  Either the visual inspection and silhouette scores are misleading, or the seemingly better solution found by the mean fusion model is not at a point where it can be exploited to achieve better EER.

\section{Conclusion}
We trained several fusion architectures to solve a verification task under uni-modal, mixed-modal, cross-modal, and full-modal regimes, and showed empirically that an architecture that combines the two modalities through averaging at the output, i.e. mean fusion, creates more discriminative visual embeddings than ones which share weights for both modalities, while still being able to perform missing-modality verification.  We also observe that it provides greater flexibility to the model, allowing solutions that place the audio and visual embedding in very different parts of the embedding space. Finally, we replicate the findings of \cite{sari_multi-view_2021} that indicate mixed-modal verification provides no added benefit over uni-modal verification.

\bibliographystyle{IEEEtran}
\bibliography{references, mm_diarization}

\begin{thebibliography}{10}
\providecommand{\url}[1]{#1}
\csname url@samestyle\endcsname
\providecommand{\newblock}{\relax}
\providecommand{\bibinfo}[2]{#2}
\providecommand{\BIBentrySTDinterwordspacing}{\spaceskip=0pt\relax}
\providecommand{\BIBentryALTinterwordstretchfactor}{4}
\providecommand{\BIBentryALTinterwordspacing}{\spaceskip=\fontdimen2\font plus
\BIBentryALTinterwordstretchfactor\fontdimen3\font minus
  \fontdimen4\font\relax}
\providecommand{\BIBforeignlanguage}[2]{{%
\expandafter\ifx\csname l@#1\endcsname\relax
\typeout{** WARNING: IEEEtran.bst: No hyphenation pattern has been}%
\typeout{** loaded for the language `#1'. Using the pattern for}%
\typeout{** the default language instead.}%
\else
\language=\csname l@#1\endcsname
\fi
#2}}
\providecommand{\BIBdecl}{\relax}
\BIBdecl

\bibitem{wan_generalized_2020}
\BIBentryALTinterwordspacing
L.~Wan, Q.~Wang, A.~Papir, and I.~L. Moreno, ``Generalized {End}-to-{End}
  {Loss} for {Speaker} {Verification},'' \emph{arXiv:1710.10467 [cs, eess,
  stat]}, Nov. 2020, arXiv: 1710.10467. [Online]. Available:
  \url{http://arxiv.org/abs/1710.10467}
\BIBentrySTDinterwordspacing

\bibitem{wang_speaker_2018}
\BIBentryALTinterwordspacing
Q.~Wang, C.~Downey, L.~Wan, P.~A. Mansfield, and I.~L. Moreno, ``Speaker
  {Diarization} with {LSTM},'' \emph{arXiv:1710.10468 [cs, eess, stat]}, Dec.
  2018, arXiv: 1710.10468. [Online]. Available:
  \url{http://arxiv.org/abs/1710.10468}
\BIBentrySTDinterwordspacing

\bibitem{fan_exploring_2021}
\BIBentryALTinterwordspacing
Z.~Fan, M.~Li, S.~Zhou, and B.~Xu, ``Exploring wav2vec 2.0 on speaker
  verification and language identification,'' \emph{arXiv:2012.06185 [cs,
  eess]}, Jan. 2021, arXiv: 2012.06185 version: 2. [Online]. Available:
  \url{http://arxiv.org/abs/2012.06185}
\BIBentrySTDinterwordspacing

\bibitem{mehdipour_ghazi_comprehensive_2016}
\BIBentryALTinterwordspacing
M.~Mehdipour~Ghazi and H.~Kemal~Ekenel, ``A {Comprehensive} {Analysis} of
  {Deep} {Learning} {Based} {Representation} for {Face} {Recognition},'' 2016,
  pp. 34--41. [Online]. Available:
  \url{https://www.cv-foundation.org/openaccess/content_cvpr_2016_workshops/w4/html/Ghazi_A_Comprehensive_Analysis_CVPR_2016_paper.html}
\BIBentrySTDinterwordspacing

\bibitem{deng_arcface_2019}
\BIBentryALTinterwordspacing
J.~Deng, J.~Guo, N.~Xue, and S.~Zafeiriou, ``{ArcFace}: {Additive} {Angular}
  {Margin} {Loss} for {Deep} {Face} {Recognition},'' \emph{arXiv:1801.07698
  [cs]}, Feb. 2019, arXiv: 1801.07698. [Online]. Available:
  \url{http://arxiv.org/abs/1801.07698}
\BIBentrySTDinterwordspacing

\bibitem{baevski_wav2vec_2020}
\BIBentryALTinterwordspacing
A.~Baevski, H.~Zhou, A.~Mohamed, and M.~Auli, ``wav2vec 2.0: {A} {Framework}
  for {Self}-{Supervised} {Learning} of {Speech} {Representations},''
  \emph{arXiv:2006.11477 [cs, eess]}, Oct. 2020, arXiv: 2006.11477. [Online].
  Available: \url{http://arxiv.org/abs/2006.11477}
\BIBentrySTDinterwordspacing

\bibitem{liu_mockingjay_2020}
\BIBentryALTinterwordspacing
A.~T. Liu, S.-w. Yang, P.-H. Chi, P.-c. Hsu, and H.-y. Lee, ``Mockingjay:
  {Unsupervised} {Speech} {Representation} {Learning} with {Deep}
  {Bidirectional} {Transformer} {Encoders},'' \emph{ICASSP 2020 - 2020 IEEE
  International Conference on Acoustics, Speech and Signal Processing
  (ICASSP)}, pp. 6419--6423, May 2020, arXiv: 1910.12638 version: 2. [Online].
  Available: \url{http://arxiv.org/abs/1910.12638}
\BIBentrySTDinterwordspacing

\bibitem{chung_out_2017}
J.~S. Chung and A.~Zisserman, ``\BIBforeignlanguage{en}{Out of {Time}:
  {Automated} {Lip} {Sync} in the {Wild}},'' in
  \emph{\BIBforeignlanguage{en}{Computer {Vision} – {ACCV} 2016
  {Workshops}}}, ser. Lecture {Notes} in {Computer} {Science}, C.-S. Chen,
  J.~Lu, and K.-K. Ma, Eds.\hskip 1em plus 0.5em minus 0.4em\relax Cham:
  Springer International Publishing, 2017, pp. 251--263.

\bibitem{shon_noise-tolerant_2019}
S.~Shon, T.-H. Oh, and J.~Glass, ``Noise-tolerant {Audio}-visual {Online}
  {Person} {Verification} {Using} an {Attention}-based {Neural} {Network}
  {Fusion},'' in \emph{{ICASSP} 2019 - 2019 {IEEE} {International} {Conference}
  on {Acoustics}, {Speech} and {Signal} {Processing} ({ICASSP})}, May 2019, pp.
  3995--3999, iSSN: 2379-190X.

\bibitem{chen_multi-modality_2020}
\BIBentryALTinterwordspacing
Z.~Chen, S.~Wang, and Y.~Qian, ``\BIBforeignlanguage{en}{Multi-{Modality}
  {Matters}: {A} {Performance} {Leap} on {VoxCeleb}},'' in
  \emph{\BIBforeignlanguage{en}{Interspeech 2020}}.\hskip 1em plus 0.5em minus
  0.4em\relax ISCA, Oct. 2020, pp. 2252--2256. [Online]. Available:
  \url{http://www.isca-speech.org/archive/Interspeech_2020/abstracts/2229.html}
\BIBentrySTDinterwordspacing

\bibitem{kang_multimodal_2020}
W.~Kang, B.~C. Roy, and W.~Chow, ``Multimodal {Speaker} {Diarization} of
  {Real}-{World} {Meetings} {Using} {D}-{Vectors} {With} {Spatial}
  {Features},'' in \emph{{ICASSP} 2020 - 2020 {IEEE} {International}
  {Conference} on {Acoustics}, {Speech} and {Signal} {Processing} ({ICASSP})},
  May 2020, pp. 6509--6513, iSSN: 2379-190X.

\bibitem{chung_who_2019}
\BIBentryALTinterwordspacing
J.~S. Chung, B.-J. Lee, and I.~Han, ``Who said that?: {Audio}-visual speaker
  diarisation of real-world meetings,'' \emph{arXiv:1906.10042 [cs, eess]},
  Jun. 2019, arXiv: 1906.10042. [Online]. Available:
  \url{http://arxiv.org/abs/1906.10042}
\BIBentrySTDinterwordspacing

\bibitem{feichtenhofer_convolutional_2016}
\BIBentryALTinterwordspacing
C.~Feichtenhofer, A.~Pinz, and A.~Zisserman, ``Convolutional {Two}-{Stream}
  {Network} {Fusion} for {Video} {Action} {Recognition},''
  \emph{arXiv:1604.06573 [cs]}, Sep. 2016, arXiv: 1604.06573. [Online].
  Available: \url{http://arxiv.org/abs/1604.06573}
\BIBentrySTDinterwordspacing

\bibitem{carreira_quo_2018}
\BIBentryALTinterwordspacing
J.~Carreira and A.~Zisserman, ``Quo {Vadis}, {Action} {Recognition}? {A} {New}
  {Model} and the {Kinetics} {Dataset},'' \emph{arXiv:1705.07750 [cs]}, Feb.
  2018, arXiv: 1705.07750. [Online]. Available:
  \url{http://arxiv.org/abs/1705.07750}
\BIBentrySTDinterwordspacing

\bibitem{jaegle_perceiver_2021}
\BIBentryALTinterwordspacing
A.~Jaegle, F.~Gimeno, A.~Brock, A.~Zisserman, O.~Vinyals, and J.~Carreira,
  ``\BIBforeignlanguage{en}{Perceiver: {General} {Perception} with {Iterative}
  {Attention}},'' \emph{\BIBforeignlanguage{en}{arXiv:2103.03206 [cs, eess]}},
  Jun. 2021, arXiv: 2103.03206. [Online]. Available:
  \url{http://arxiv.org/abs/2103.03206}
\BIBentrySTDinterwordspacing

\bibitem{afouras_deep_2018}
T.~Afouras, J.~S. Chung, A.~Senior, O.~Vinyals, and A.~Zisserman, ``Deep
  {Audio}-visual {Speech} {Recognition},'' \emph{IEEE Transactions on Pattern
  Analysis and Machine Intelligence}, pp. 1--1, 2018, conference Name: IEEE
  Transactions on Pattern Analysis and Machine Intelligence.

\bibitem{petridis_end--end_2017}
\BIBentryALTinterwordspacing
S.~Petridis, Y.~Wang, Z.~Li, and M.~Pantic, ``End-to-{End} {Audiovisual}
  {Fusion} with {LSTMs},'' \emph{arXiv:1709.04343 [cs]}, Sep. 2017, arXiv:
  1709.04343. [Online]. Available: \url{http://arxiv.org/abs/1709.04343}
\BIBentrySTDinterwordspacing

\bibitem{sari_multi-view_2021}
L.~Sarı, K.~Singh, J.~Zhou, L.~Torresani, N.~Singhal, and Y.~Saraf, ``A
  {Multi}-{View} {Approach} to {Audio}-{Visual} {Speaker} {Verification},'' in
  \emph{{ICASSP} 2021 - 2021 {IEEE} {International} {Conference} on
  {Acoustics}, {Speech} and {Signal} {Processing} ({ICASSP})}, Jun. 2021, pp.
  6194--6198, iSSN: 2379-190X.

\bibitem{wang_what_2020}
\BIBentryALTinterwordspacing
W.~Wang, D.~Tran, and M.~Feiszli, ``What {Makes} {Training} {Multi}-{Modal}
  {Classification} {Networks} {Hard}?'' \emph{arXiv:1905.12681 [cs]}, Apr.
  2020, arXiv: 1905.12681. [Online]. Available:
  \url{http://arxiv.org/abs/1905.12681}
\BIBentrySTDinterwordspacing

\bibitem{sandler_mobilenetv2_2019}
\BIBentryALTinterwordspacing
M.~Sandler, A.~Howard, M.~Zhu, A.~Zhmoginov, and L.-C. Chen, ``{MobileNetV2}:
  {Inverted} {Residuals} and {Linear} {Bottlenecks},'' \emph{arXiv:1801.04381
  [cs]}, Mar. 2019, arXiv: 1801.04381. [Online]. Available:
  \url{http://arxiv.org/abs/1801.04381}
\BIBentrySTDinterwordspacing

\bibitem{he_deep_2015}
\BIBentryALTinterwordspacing
K.~He, X.~Zhang, S.~Ren, and J.~Sun, ``Deep {Residual} {Learning} for {Image}
  {Recognition},'' \emph{arXiv:1512.03385 [cs]}, Dec. 2015, arXiv: 1512.03385.
  [Online]. Available: \url{http://arxiv.org/abs/1512.03385}
\BIBentrySTDinterwordspacing

\bibitem{nagrani_voxceleb_2020}
\BIBentryALTinterwordspacing
A.~Nagrani, J.~S. Chung, W.~Xie, and A.~Zisserman,
  ``\BIBforeignlanguage{en}{Voxceleb: {Large}-scale speaker verification in the
  wild},'' \emph{\BIBforeignlanguage{en}{Computer Speech \& Language}},
  vol.~60, p. 101027, Mar. 2020. [Online]. Available:
  \url{https://www.sciencedirect.com/science/article/pii/S0885230819302712}
\BIBentrySTDinterwordspacing

\bibitem{nagrani_voxceleb_2017}
\BIBentryALTinterwordspacing
A.~Nagrani, J.~S. Chung, and A.~Zisserman, ``{VoxCeleb}: a large-scale speaker
  identification dataset,'' \emph{Interspeech 2017}, pp. 2616--2620, Aug. 2017,
  arXiv: 1706.08612. [Online]. Available: \url{http://arxiv.org/abs/1706.08612}
\BIBentrySTDinterwordspacing

\bibitem{de_paz_centeno_mtcnnlicense_nodate}
\BIBentryALTinterwordspacing
I.~de~Paz~Centeno, ``\BIBforeignlanguage{en}{mtcnn/{LICENSE} at master ·
  ipazc/mtcnn}.'' [Online]. Available: \url{https://github.com/ipazc/mtcnn}
\BIBentrySTDinterwordspacing

\bibitem{zhang_joint_2016}
\BIBentryALTinterwordspacing
K.~Zhang, Z.~Zhang, Z.~Li, and Y.~Qiao, ``Joint {Face} {Detection} and
  {Alignment} using {Multi}-task {Cascaded} {Convolutional} {Networks},''
  \emph{IEEE Signal Processing Letters}, vol.~23, no.~10, pp. 1499--1503, Oct.
  2016, arXiv: 1604.02878. [Online]. Available:
  \url{http://arxiv.org/abs/1604.02878}
\BIBentrySTDinterwordspacing

\bibitem{malli_keras-vggface_2022}
\BIBentryALTinterwordspacing
R.~C. Malli, ``keras-vggface,'' May 2022, original-date: 2016-10-17T15:07:40Z.
  [Online]. Available: \url{https://github.com/rcmalli/keras-vggface}
\BIBentrySTDinterwordspacing

\bibitem{mcinnes_umap_2020}
\BIBentryALTinterwordspacing
L.~McInnes, J.~Healy, and J.~Melville, ``{UMAP}: {Uniform} {Manifold}
  {Approximation} and {Projection} for {Dimension} {Reduction},''
  \emph{arXiv:1802.03426 [cs, stat]}, Sep. 2020, arXiv: 1802.03426. [Online].
  Available: \url{http://arxiv.org/abs/1802.03426}
\BIBentrySTDinterwordspacing

\bibitem{adamw}
\BIBentryALTinterwordspacing
I.~Loshchilov and F.~Hutter, ``Fixing weight decay regularization in adam,''
  \emph{CoRR}, vol. abs/1711.05101, 2017. [Online]. Available:
  \url{http://arxiv.org/abs/1711.05101}
\BIBentrySTDinterwordspacing

\bibitem{falcon2019pytorch}
W.~Falcon, ``Pytorch lightning,'' \emph{GitHub. Note:
  https://github.com/PyTorchLightning/pytorch-lightning Cited by}, vol.~3,
  2019.

\end{thebibliography}

\appendices

\section{Model Architecture}

\subsection{Backbones}

The visual backbone follows the ResNet v1.5 architecture.  This model is ResNet-50 as described in \cite{he_deep_2015}, except that for bottleneck blocks that require downsampling the downsampling (stride = 2) is done in the 3x3 convolutional layer.  The batch normalization layers after all convolutional layers are all instantiated with pytorch's nn.BatchNorm2d class with parameters:

\vspace{3mm}

\begin{description}
   \item[num\_features] output dim of previous layer
   \item[eps] 1e-05
   \item[momentum] 0.1
   \item[affine] True
   \item[track\_running\_state] True
\end{description}

\vspace{3mm}

We average pool the final 2d feature map over all channels to go from (B x T x C x H x W) to (B x T x C x 1).  Where B, T, C, H, W are the batch size, number of frames, channels, feature map height and feature map width.  Finally, we average over the T frames to get a final feature representation of dimension 2048.

The audio backbone uses a stack of inverted residual blocks as described in  \cite{sandler_mobilenetv2_2019}.  We have a stack of 12 such blocks with inverted residual settings identical to \cite{sari_multi-view_2021}.  All convolutional layers are followed by batch normalization identical to the video backbone and a ReLU6 activation.  At the input, we begin with a 1d convolution on the spectrogram map to increase the channel dimension to 32 before feeding through the 12 residual blocks.  At the output, we apply another 1d convolution to bring the channel dimension from 256 to 356.  At this point the tensor dimensions are 356 x 4 x 10, where the second and third dimensions started as the number of Mel bins and number of time steps in the Mel spectrogram.  We average the middle dimension to get a 356 x 10 tensor and then transpose and apply two linear projections with weight matrices of dimension 356 x 356 and 356 x 1, each followed by a leaky ReLU activation.  This gives us a tensor of dimension 10 x 1, which is then right multiplied with the 356 x 10 tensor to give us a 356 dimensional output (after squeezing the hanging dimension).

\subsection{Fusion}

\begin{figure*}[ht]
    \centering
    \includegraphics[width=\textwidth]{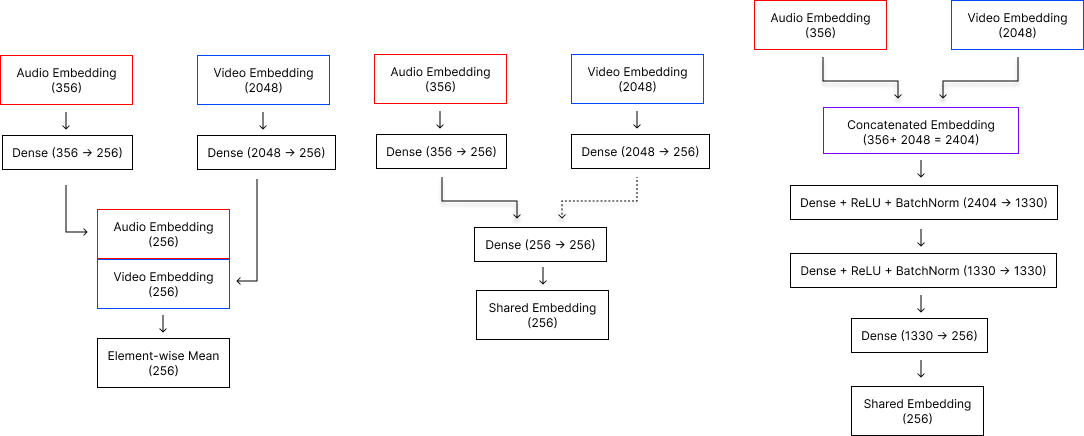}
    \caption{Fusion architectures.  (Left) Mean fusion.  (Middle) Multiview fusion, dotted line indicates that only one input from the audio or visual stream is passed through at a time.  (Right) Multi-layer perceptron fusion.}
    \label{fig:fusion_arch}
\end{figure*}

Figure \ref{fig:fusion_arch} shows each of the three fusion architectures.  All fusion architectures take the 2048 and 356 dimensional video and audio embeddings from the model backbones as input.  In the mean fusion and MLP architectures, during training, dropout with probability 0.1 is applied to the two embeddings before being fed into the fusion module.

The mean fusion architecture uses two standard linear layers with bias to map each embedding to dimension 256.  The embeddings are then averaged over all feature dimensions.

The MLP fusion architecture concatenates the audio and video embeddings to form an input of size 2048 + 356 = 2404.  This is then passed through 3 linear layers each followed by a leaky ReLU activation and batch normalization identical to that described in the visual backbone.  During training, dropout with probability 0.1 is also applied after the batch normalization of the first two layers.  In our setup, the layers were instantiated with pytorch's nn.Linear with parameters as follows:

\vspace{3mm}

\begin{description}
   \item[linear\_1] in\_features=2404, out\_features=1330, bias=True
   \item[linear\_2] in\_features=1330, out\_features=1330, bias=True
   \item[linear\_3] in\_features=1330, out\_features=256, bias=True
\end{description}

\vspace{3mm}

The multi-view fusion begins with linear projections identical to the mean fusion architecture which map the audio and video embeddings to a 256 dimensional space.  These vectors are both separately sent through the shared `classification' layer, a linear layer with bias that maps to the same dimensionality (256).  The classification layer is followed by a ReLU layer and, during training, a dropout layer with dropout probability 0.1.  

\section{Training Details}

\subsection{Loss Function}

We use the loss function described in \cite{deng_arcface_2019} section 2.  Our weight matrix given the parameters below is $256 \times 5894$.

\vspace{3mm}

\begin{description}
   \item[input features]: 256
   \item[output features]: 5894
   \item[feature scale]: 16
   \item[margin]: 0.125
\end{description}

\vspace{3mm}

\subsection{Optimizer}

We use the AdamW \cite{adamw} optimizer implemented with pytorch's torch.optim.AdamW class with the following parameters set:

\vspace{3mm}

\begin{description}
   \item[lr (learning rate)]: 0.001
   \item[betas]: 0.9, 0.999
   \item[eps]: 1.0e-08
   \item[weight\_decay]: 0.01
   \item[amsgrad]: False
\end{description}

\subsection{Data Processing}

\subsubsection{Random Masking}

During training for the mean fusion and MLP architectures we randomly mask (replace with zeros) the audio or video features at the output of the model backbones.  Specifically, with probability $\frac{1}{3}$ each, we do \textbf{one} of the following:
\begin{itemize}
    \item mask the 2048 dimensional video backbone output
    \item mask the 356 dimensional audio backbone output
    \item do not mask
\end{itemize}

\subsubsection{Audio transforms}

Audio is sampled at 16000 hz, clipped to 1.5 seconds of audio, and then transformed to log Mel-spectrograms with 64 Mel bins, 512 fast Fourier transforms, windows length of 400 samples, and hop length of 160 samples.  Each mel bin is then shifted and scaled by its mean and standard deviation across all time windows for all clips in the Voxceleb2 training set.

\subsubsection{Video transforms}

Video is first downsampled to 2 fps and clipped to 1.5 seconds (3 frames).  The video frames are resized to 112 x 112 and are pixel-wise shifted and scaled with mean values [0.4582268298, 0.3447833359, 0.3283427358] and scaling values [0.2709922791, 0.2274252474, 0.2343513072] computed from the average and standard deviation of all RGB pixels from the Voxceleb2 train set.

\subsection{Misc Training Parameters}

The following are set during training:

\begin{itemize}
    \item Use random seed 4242 passed to pytorch-lightning's \cite{falcon2019pytorch} seed\_everything function.
    \item Gradient clipping with max l2 norm 5.0
    \item 32 bit precision
    \item batch size 128
\end{itemize}

\section{Missing Data}

The following clips in the Voxceleb1 test set were missing.

\begin{table}[h]
    \centering
    \begin{tabular}{c|c|c}
         Video ID & Speaker ID & Track ID \\
         	Cb07--9j3-Q	& id10307 &	1 \\
        GyJHpVQmcvc	& id10309	& 2 \\
        HTL8iLI75TY	& id10304	& 4 \\
        RDFdX3VxjUQ	& id10281	& 5 \\
        RNYNkXzY5Hk	& id10284	& 18 \\
        RNYNkXzY5Hk	& id10284	& 20 \\
        ZLzkvnq0JxI	& id10305	& 3 \\
        ZNT2uhs3jF4	& id10307	& 1 \\
        arklnCzCq48	& id10283	& 2 \\
        arklnCzCq48	& id10283	& 3 \\
        f8Ms66atECE	& id10307	& 7 \\
        gegIAYxfpVA	& id10293	& 5 \\
        gegIAYxfpVA	& id10293	& 6 \\
        nJbBcMdxQU4	& id10305	& 17 \\
        p5g1heXp34o	& id10283	& 3 \\
        p5g1heXp34o	& id10283	& 8 \\
        vv4mvANXHcs	& id10283	& 2 \\
        vv4mvANXHcs	& id10283	& 3 \\
         & 
    \end{tabular}
    \label{tab:missing videos}
\end{table}

\end{document}